\documentclass[10pt, a4paper]{article}
\usepackage{lrec}
\usepackage{graphicx}
\usepackage{tabularx}
\usepackage{soul}
\usepackage{csvsimple}
\usepackage{subcaption}

\usepackage{hyperref}
\usepackage{xstring}
\usepackage{url}
\usepackage{booktabs}
\usepackage{multirow}
\usepackage{subcaption}
\usepackage{longtable}
\usepackage{siunitx}
\usepackage{color}
\usepackage{ifthen}
\usepackage{booktabs}

\definecolor{darkgreen}{rgb}{0.09, 0.45, 0.27}
\definecolor{pink}{rgb}{0.78, 0.08, 0.52}

\title{\texttt{CCNet}: Extracting High Quality Monolingual Datasets from Web Crawl Data}

\name{Guillaume Wenzek$^*$, 
  Marie-Anne Lachaux$^*$, Alexis Conneau, Vishrav Chaudhary, \\
  \textbf{\large Francisco Guzm\'an, Armand Joulin, Edouard Grave}}

\address{Facebook AI \\
  \{\texttt{guw, malachaux, aconneau, vishrav, fguzman, ajoulin, egrave\}@fb.com}}

\abstract{
Pre-training text representations have led to significant improvements in many areas of natural language processing.
The quality of these models benefits greatly from the size of the pretraining corpora as long as its quality is preserved. 
In this paper, we describe an automatic pipeline to extract massive high-quality monolingual datasets from Common Crawl for a variety of languages.
Our pipeline follows the data processing introduced in fastText~\cite{mikolov2017advances,grave2018learning}, that deduplicates documents and identifies their language. 
We augment this pipeline with a filtering step to select documents that are close to high quality corpora like Wikipedia.
\newline 
}

\begin{document}

\maketitleabstract

\section{Introduction}

Pre-trained text representations have brought significant performance gains on many natural language processing tasks~\cite{peters2018deep}.
Since the introduction of Transformers~\cite{vaswani2017attention} and BERT~\cite{devlin2018bert}, we have a seen a steady improvement in the quality of these pre-trained models,
mainly driven by increasing the size of the pre-training corpora~\cite{radford2019language,yang2019xlnet,lan2019albert}.
Nonetheless, the size only does not guarantee better models and  the quality of the data has to be preserved, 
which has lead to the use of \emph{ad-hoc} datasets created by concatenating existing high-quality data sources like Wikipedia.
Unfortunately, such datasets cannot be replicated as easily for low-resources languages, as many have much smaller curated datasets such as Wikipedia.

In this paper, we present a data collection pipeline that allows to gather massive monolingual corpora of high quality in a variety of languages, including many low-resource ones.
The principles of our pipeline are general and we show the results of its application to data collected by the Common Crawl project.\footnote{\url{https://commoncrawl.org/about/}}
Common Crawl is a massive non-curated dataset of webpages in many languages, mixed together in temporal snapshots of the web.
Our pipeline performs standard document deduplication and language identification similar to \newcite{grave2018learning}, but differs in two ways:
first, we preserve the document-level structure to allow for the training of paragraph-level representations like BERT~\cite{devlin2018bert} ;
second, we add an optional monolingual filtering step that selects documents that are close to high quality sources, like Wikipedia.
This is achieved by training a language model on the targeted sources and use the perplexity as a scoring function for documents.
Our pipeline can be applied to any number of Common Crawl snapshots and takes 8.5 hours to process per snapshot on 5000 CPU cores.
For example, the dataset obtained by pre-processing the February 2019 snapshot is composed of 1.5 billions documents in 174 languages. 
There are 700 millions filtered documents in English alone, corresponding to 532 billions tokens.
That is 120 times bigger than the data used in \newcite{devlin2018bert}.

This paper is organized as follows: we first present the Common Crawl corpora, followed by our overall pipeline to filter high quality documents from it. 
We then describe additional tools that can be used to tailor the filtering to a targeted corpora.
Finally, we give in depth statistics about the dataset obtained from pre-processing a single Common Crawl snapshot.
The pipeline and the tools are publicly available\footnote{\url{github.com/facebookresearch/cc_net}}.

\section{Related work}

Preprocessing of massive datasets for training text representations has been developed in the context of word embeddings, such as word2vec~\cite{mikolov2013distributed},  GloVe~\cite{pennington2014glove} or fastText~\cite{mikolov2017advances}.
In particular, our pipeline follows the fastText pipeline of \newcite{grave2018learning} where Common Crawl is split into monolingual datasets using a language identifier based on fastText~\cite{joulin2016fasttext}.

Common Crawl has been used in the context of language modeling to evaluate ~$n$-gram statistics~\cite{buck2014n}. 
More recently, \newcite{baevski2019cloze} pre-trained a BERT-like model on  Common Crawl as preprocessed in~\newcite{grave2018learning}. 
In general, progress in sentence representations has been observed by increasing the size of the pre-training corpora~\cite{yang2019xlnet,liu2019roberta,raffel2019exploring}.
In particular, and concurrently to our work, \newcite{raffel2019exploring} used a large scale dataset based on Common Crawl to train text representations.
Existing work using web based datasets have been using English specific preprocessing, such as keeping URLs shared on Reddit or using hand-crafted filtering rules.
As opposed to these approaches, our pipeline can easily be applied to many languages other than English.
Closer to this work,~\newcite{suarez2019asynchronous} has improved the pipeline of~\newcite{grave2018learning}, showing that large monolingual corpora can be extracted from Common Crawl rapidly even with limited resources. Our work follows a similar pipeline with an additional step to select high-quality documents. 

\begin{figure*}[t!]
  \center
  \includegraphics[width=0.8 \textwidth]{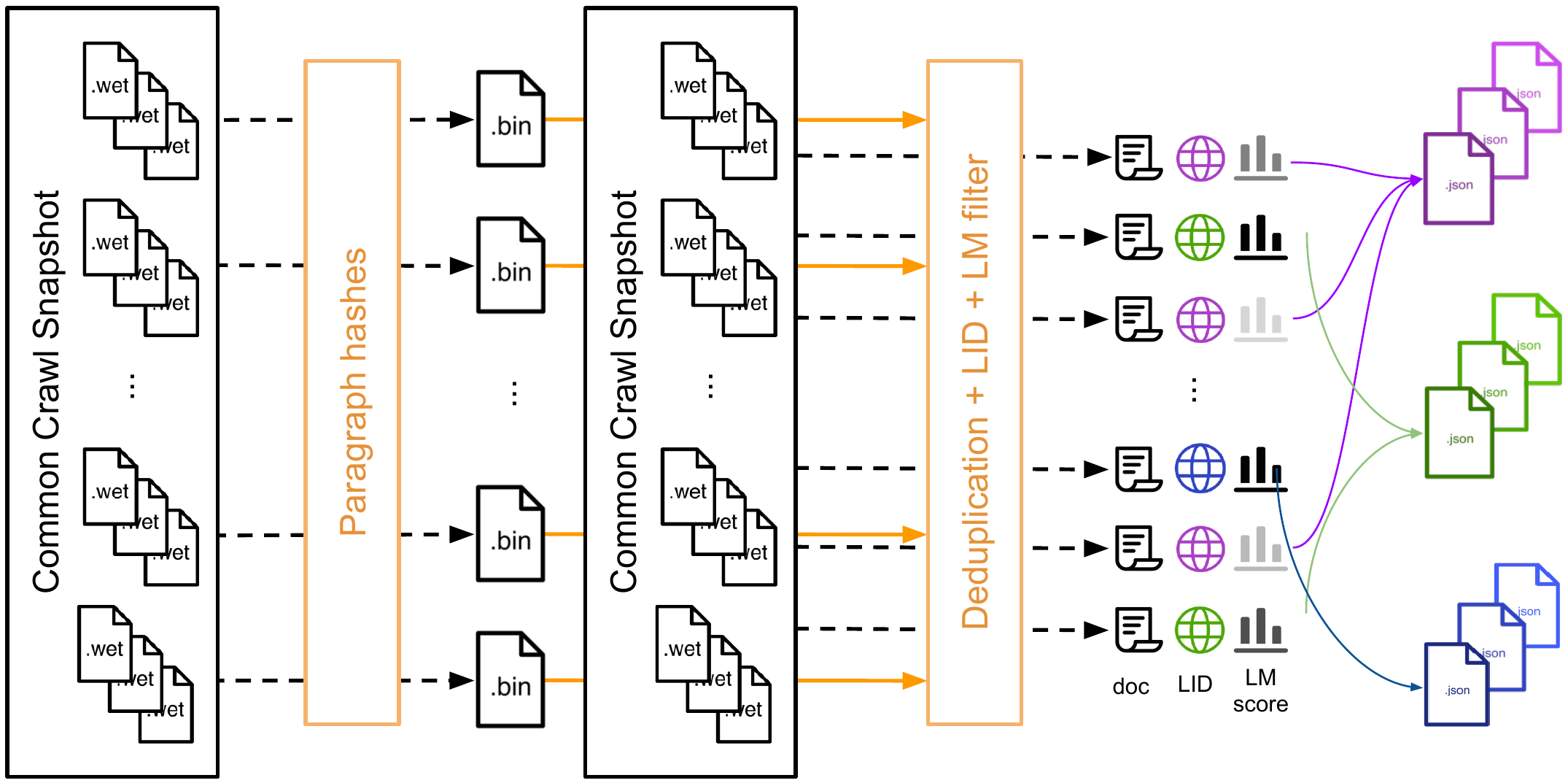}
  \caption{We show the whole pipeline for downloading and processing one snapshot of Common Crawl. First we download all the wet files and compute the paragraph hashes that we group and save into binary files. Then we process every document of the wet files independently: we deduplicate the paragraph using the binary files, we do a language identification and compute language model perplexity score. Finally, we regroup the documents into json files by language and perplexity score. The steps of the pipeline indicated with dashed arrows are parallelisable.  }
  \label{figure:pipeline}
\end{figure*}

\section{Methodology}


Every month, Common Crawl releases a snapshot of the web obtained by randomly exploring and sampling URLs. Each webpage is made available different formats: raw (WARC), UTF-8 text (WET), and meta-data (WAT).
There is little content overlap between monthly snapshots.
The complete archive consists of petabytes of data collected over $8$ years of web crawling.  
The webpages are crawled from the whole web without restriction; they come in many different languages and in the quality of the text varies greatly.  
The Common Crawl represents a rich resource for monolingual data that comprises a large variety of domains, yet poses challenges due to the large quantity of noisy text. 

Here we describe our the methodology used to fetch, deduplicate and filter the Common Crawl data. We focus on preprocessing the text (WET) format of the common crawl snapshots. 
Our pre-processing pipeline consists of several steps that we describe in this section. An overview of the pipeline is  illustrated in figure \ref{figure:pipeline}.


\subsection{Preprocessing}

Each snapshot contain between 20 and 30TB of uncompressed plain text, corresponding to approximately 3 billion web pages
(for instance the Feb. 2019 snapshot contains 24TB of data).
We download and process each snapshot independently. 
For each snapshot, we regroup WET files into shards of $5$GB each.
This makes up for 1600 shards for Feb. 2019 crawl.
These shards are saved into a JSON file where one entry corresponds to one web page.

\subsection{Deduplication}

The first step of our pipeline consists in removing duplicated paragraphs across the different web pages in a snapshot, as they represent $70\%$ of the text.
We first normalize each paragraph by lower-casing all characters, replacing numbers by a placeholder (i.e. $0$) and removing all Unicode punctuation and accent marks. 

Then, the deduplication is done in two independent steps.
First, for every shard, we compute a hash code for each paragraph and save them into a binary file. 
We use the first 64-bits of SHA-1 digits of the normalized paragraphs as the key.
Then, we deduplicate every shard by comparing it with either $1$, a subset or all of the binary files.

The impact of this choice is discussed in \ref{section:discussion}
These steps are independent for each shard and can thus be distributed.
In addition to removing web copies, this step gets rid of a lot boilerplate such as navigation menus, cookie warnings and contact information.
In particular, it removes significant amount of English content from webpages in other languages.
This makes the language identification, which is the next step of our pipeline, more robust.

\begin{figure*}[t]
\center
  \includegraphics{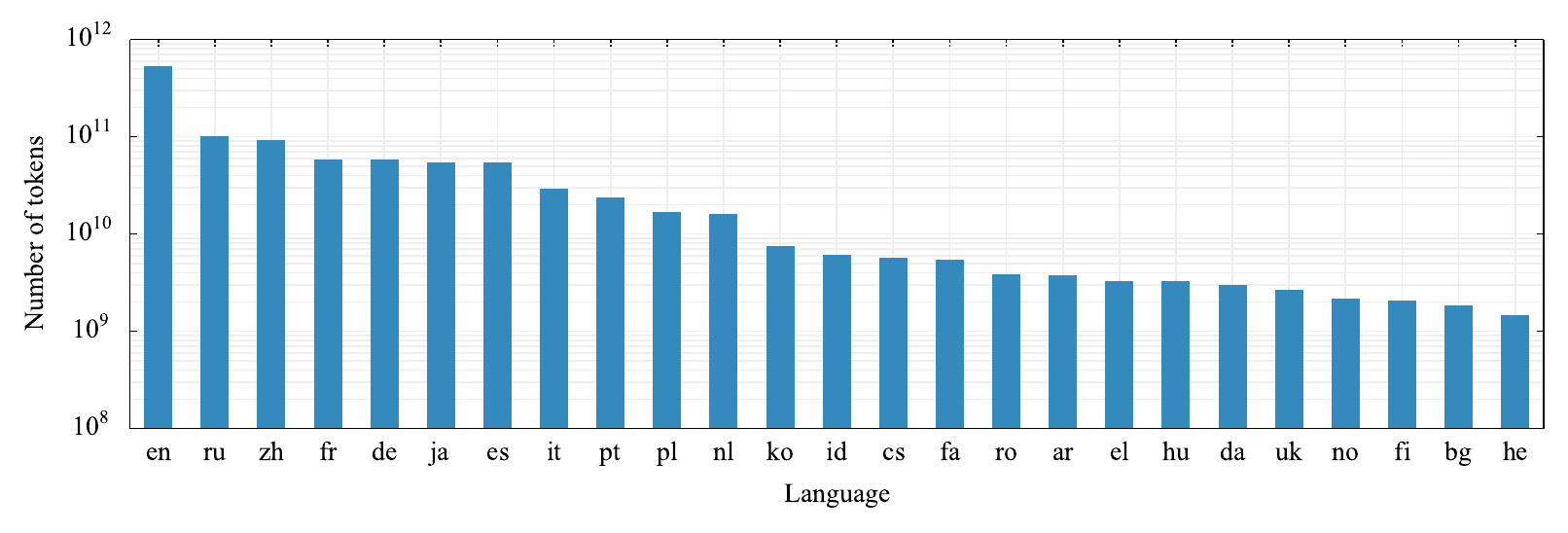}
  \caption{Number of tokens per language for the Feb. 2019 snapshot after deduplication. We display the histogram with logarithmic scale.}
  \label{figure:number_tokens}
\end{figure*}

\subsection{Language identification}
The second step of our pipeline consists in splitting data per language. 
Following~\newcite{grave2018learning}, we use the language classifier from fastText~\cite{joulin2016bag,grave2018learning}. 
The fastText language identifier was trained on Wikipedia, Tatoeba and SETimes. 
It uses characters $n$-grams as features, and the hierarchical softmax.
It supports 176 languages and outputs a score for each of them in the $[0,1]$ range. 
It processes $1$k documents per second on a single CPU core. 
For every web page we compute the most probable language, and the corresponding classifier score.
If this score is higher than 0.5, we classify the document in the corresponding language.
Otherwise, the language is not clearly identified, and we discard the corresponding page.

\subsection{LM filtering}
At this step of the pipeline, there are still documents with low quality content. 
A way to filter out these samples, is to compute a score of similarity of a web page with a targeted domain such as Wikipedia.
In this paper, we propose to use the perplexity of a language model trained on the targeted domain as the quality score. 

More precisely, for each language, we train a sentence piece tokenizer~\cite{kudo2018subword} and a language model on data from the targeted domain.
We use a $5$-gram Kneser-Ney model as implemented in the KenLM library~\cite{Heafield-kenlm} because of its efficiency to process large quantity of data. 
Then, we tokenize each page in our dataset, with our sentence piece tokenizer and compute the perplexity of each paragraph using our language model. 
The lower the perplexity, the closer the data is to the targeted domain. 
At the end of this step, each language is split into three even parts $head$, $middle$ and $tail$, corresponding to the perplexity score.
In section \ref{section:metrics} we show perplexity distributions for one snapshot of Common Crawl.

We have trained sentence piece and Kneser-Ney language models on Wikipedia for $48$ languages.
We make these models publicly available in the repository. 
We also provide code to train sentence piece and Kneser-Ney language models and compute the terciles thresholds if the user wants to use other data to filter Common Crawl.

\subsection{Reproducing results without the pipeline}

Reconstructing the dataset by running our pipeline requires a lot of resources and time.
Together with the release of the pipeline, we provide a tool to efficiently reproduce the results of this work.
This tool builds on a file containing URLs of webpages and reconstructs the final output of our pipeline from this file. 

\section{Ablation study}
\label{section:discussion}
In this section, we discuss the impact of several design choices in our pipeline on the resulting datasets.

\subsection{Order of LID and deduplication steps}
\begin{figure*}[t]
\center
  \includegraphics[scale=1.2]{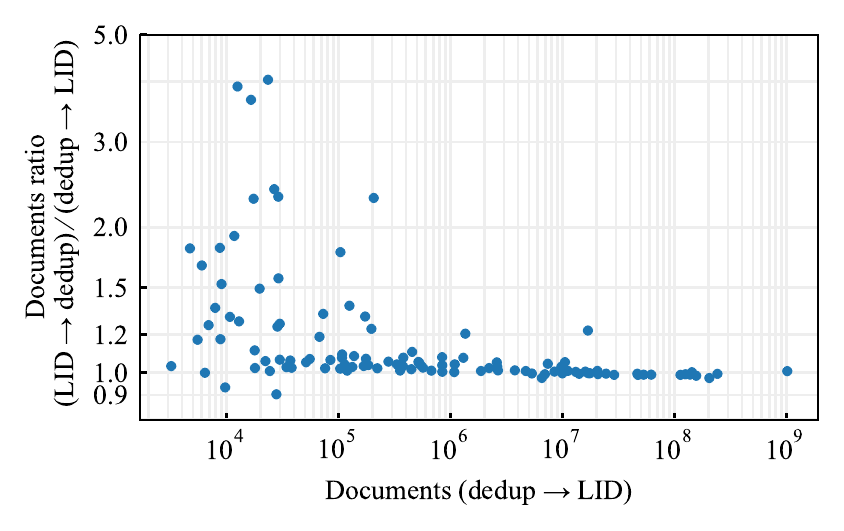}
  \caption{
  Impact of doing "Deduplication then LID" rather than "LID then Deduplication".
  Y-axis shows per language-ratio of number of documents between the two methods.
  X-axis is the number of documents found for each language using LID scores obtained after deduplication.
  Low resources languages benefits the more from doing "Deduplication then LID" 
  Stats estimated on 1\% of Feb. 2019 snapshot.
  }
  \label{figure: lid_exp}
\end{figure*}

Contrarily to \cite{grave2018learning}, we have chosen to deduplicate the data before language identification, because a lot of English boilerplate, such as cookie warnings, is present in pages of other languages.
A significant amount of this noisy data is removed by deduplication which allows for better language identification.
This is particularly important for some low resource languages.
In Figure~\ref{figure: lid_exp} we report the relative increase in number of documents when doing "deduplication then LID" instead of "LID then deduplication".
We observe that a lot of low resource language documents were mis-classified before deduplication (generally to English), or discarded because no language could be identified.

%

\subsection{Impact of the amount of deduplication}
\begin{figure}[ht!]
  \includegraphics{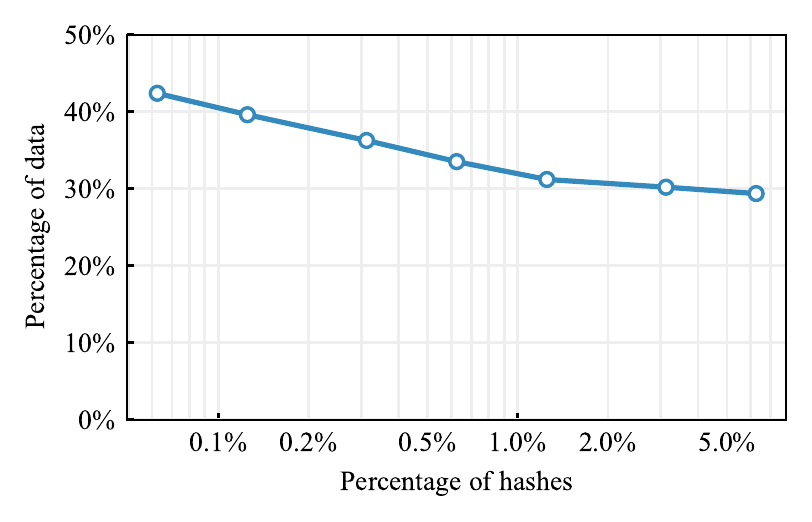}
  \caption{Amount of data remaining after deduplication with different fraction of the dataset. These statistics are computed on one shard.}
  \label{figures:dedup_shard_impact}
\end{figure}

\begin{figure}[ht!]
  \includegraphics{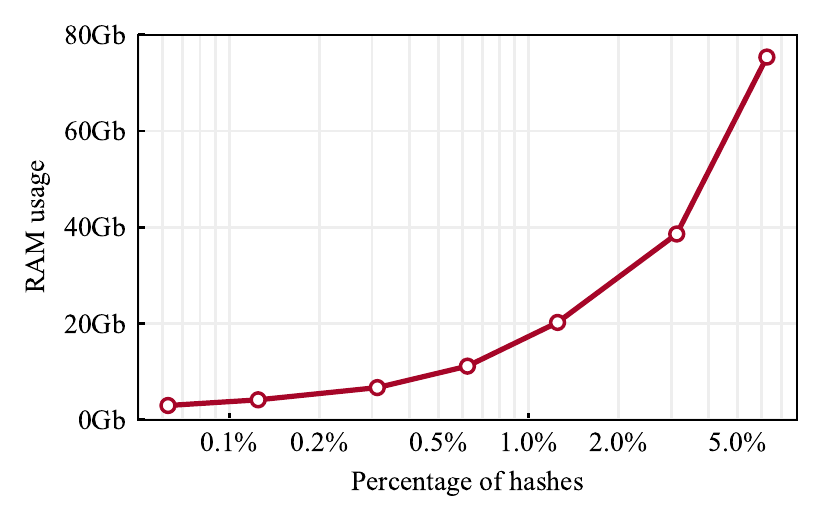}
  \caption{RAM usage when loading hashes from different fraction of the dataset. Computed on one shard.}
  \label{figures:dedup_shard_ram}
\end{figure}

For deduplication, we can compare paragraphs hashes shard by shard, across N shards or across the whole snapshot (1600 shards).
The higher N, the higher the number of documents removed and the more RAM the algorithm will use.
We show in \ref{figures:dedup_shard_impact} the amount of data remaining (percentage of number of characters) for one shard of the snapshot Feb. 2019 after deduplication across 1, 2, 5, 10, 20, 50 and 100 shards.
After deduplication across 1 shard, there is 42\% of characters remaining and 28\% across 100 shards.
Loading hashes from 50 represents 1.5B unique hashes, making up 13.5GB on disk.
Using a memory efficient hashset\footnote{\url{github.com/greg7mdp/parallel-hashmap}} we can fit those into 40GB of RAM.
In \ref{figures:dedup_shard_ram} we show how the RAM increase when we try to load more hashes in memory.
We found 50 shards to be a reasonable trade-off and are therefore running the deduplication on blocks corresponding to $3\%$ of the corpus.

\subsection{Benchmarking}

The pipeline is massively parallelizable but still has to run in two steps because of the deduplication which requires to compare billions of documents paragraphs.
In our case we chose shards of 5GB as the smallest unit of parallelisation.
One dump is divided in 1600 shards, each containing around $1.6M$ documents. 
Computing the hashes of paragraphs is done at about $600$ doc/s on one CPU core, while downloading the files at the same time.
This means that one shard of about $1.6M$ documents is done in $45$ min.
We compute all the hashes in $45$ minutes on $1600$ CPUs.
In one pass, the next step removes duplicates, and performs language identification, sentence piece tokenization, language modeling and splitting based on language.
Each shard creates 3 files for the top 48 languages for which we have a LM, and one file for each other language where we don't have a LM.
Each of those processing require a significant amount of RAM but the memory can be shared across processes since it is read only.
This step is significantly longer than the previous one.
We allocate 17 processes to one shard.
The master process is responsible for downloading the data and distributing the raw documents to the 16 workers as well as writings the results to disk.
The worker threads process around $40doc/s$, processing the whole shard in about $40$ minutes.
Removing the duplicated parapgraphs takes $40\%$ of the time.
This step is computationally less expensive than the following ones but is done on all the data, as opposed to the next steps which are only applied to the deduplicated data.
The language identifier takes $12.5\%$ of CPU time, sentence piece $33\%$ and the LM $13\%$.
Finally we regroup the files produced at the previous steps in chunks of ~5Gb.
This can be run in parallel for each output file, and since gzip archive can be concatenated without being decompressed first it's very fast and runs in matter of minutes.
The total processing time is about 9 hours using 5000 CPU cores for one snapshot.


\section{Metrics about the resulting dataset}
\label{section:metrics}

In this section, we report statistics corresponding to the corpus obtained after applying our pipeline on the Feb. 2019 snapshot of Common Crawl.

\subsection{Statistics per language}
After preprocessing it, we get $3.2$TB of compressed documents in $174$ languages. 
In table~\ref{table:table_counts_all}, we give the sizes of each monolingual corpora for the $130$ languages for which we have more than $1000$ documents.
We also compute the number of tokens and sentences for each language, and report them in Figure~\ref{figure:number_tokens}.
The tokens were obtained by using the Sentence Piece tokenizer that was used in our preprocessing pipeline. The sentences were split using Moses.
The three largest languages are English (en) with 532B tokens, Russian (ru) with 101B tokens and Chinese (zh) with 92B tokens.
We obtained 11 languages with more than 10B tokens, and 27 languages with more than 1B tokens.
In terms of documents, the three largest languages are English (en) with 706M documents, Russian (ru) with 167M and German (de) with 105M.
There are 12 languages with more than 10M documents and 29 languages containing more than 1M documents. 
Common Crawl is also a good source for lower resource languages.
For example Afrikaans (af), Gujarati (gu), Khmer (km) and Burmese (my) contains respectively 160MB, 190MB, 154MB and 440MB of data.
In comparison Wikipedia contains 103MB, 88MB, 71MB and 153MB of data for these languages.
And more resources are available through the 60 dumps of Common Crawl.
These numbers could probably be improved by increasing the recall of the LID model for low-resource languages.

\begin{figure*}[t]
\center
  \includegraphics{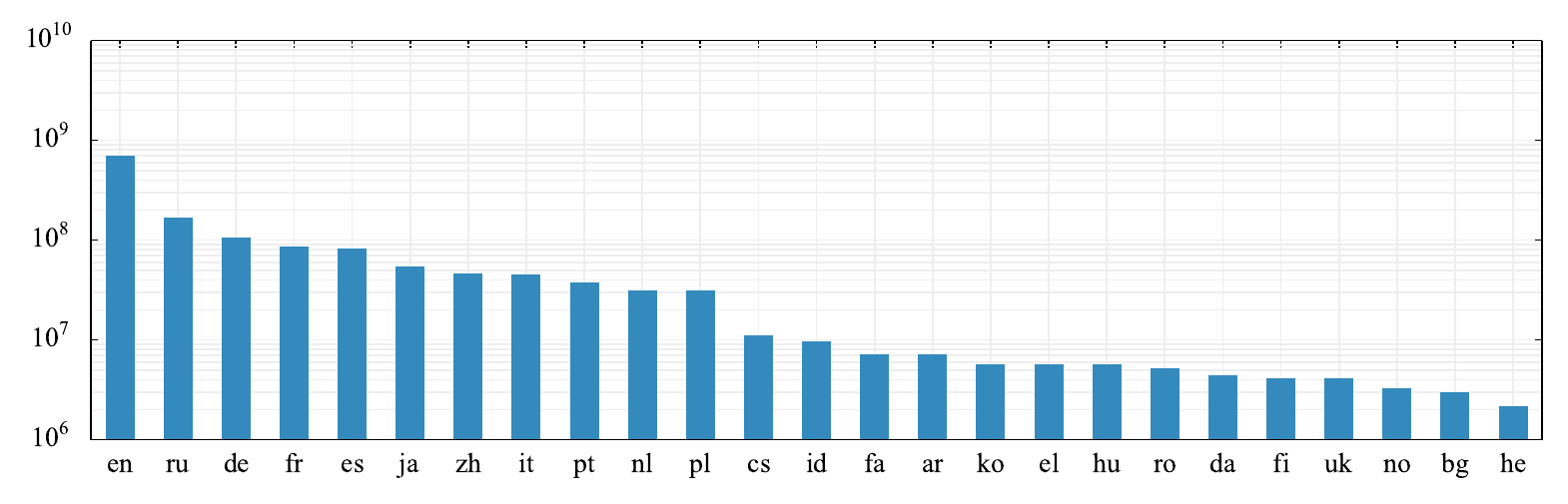}
  \caption{Number of documents per language for the Feb. 2019 snapshot after deduplication. We display the histogram with logarithmic scale. We display statistics for 25 languages only. All statisctics are available in table \ref{table:table_counts_all}} 
  \label{figure:number_documents}
\end{figure*}

\subsection{Statistics from the language model}

\begin{figure}[ht!]
  \centering
  \includegraphics[width=0.9\linewidth]{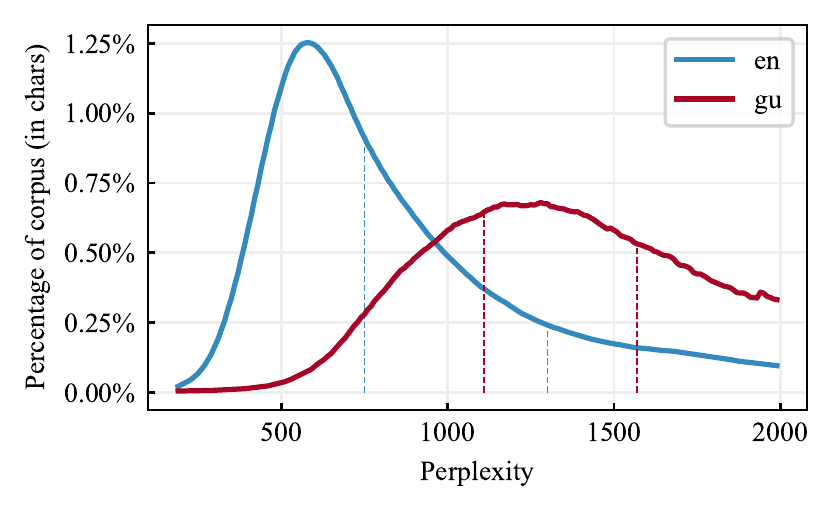}
  \caption{
    Histogram of language model perplexities for the Feb. 2019 Common Crawl snapshot.
    The two histograms correspond to English, which is the largest dataset, and Gujarati which is a low-resource language.
    Vertical lines correspond to perplexity thresholds applied to split the corpus in \textit{head}/\textit{middle}/\textit{tail}.
  }
  \label{figure:perplexity}
\end{figure}

We found that perplexity was a relative good proxy for quality. Journalistic and well written content ends up in the \textit{head} of our dataset. 
Some documents which contained a lot of keywords list passes through deduplication and LID but receive a high perplexity.
Some documents despite being valid text ends up in the \textit{tail} because they have a vocabulary very different from  Wikipedia.
This includes blog comments with spoken-like text, or very specialized forums with specific jargon.
We decided to not remove content based on the LM score because we think that some of it could be useful for specific applications.

Some languages have very spiked distribution of perplexity while others are more spread out.
We postulate that this is rather due to the variance in the Wikipedia sizes used for training the LM than to some language having less high-quality content.
Therefore we decided to use different perplexity thresholds for each language.
The thresholds have been picked to split the corpus in 3 parts of equal size.
In Figure~\ref{figure:perplexity} we show the perplexity distribution for two languages English and Gujarati using their respective LM.
English LM was trained on 534M of text while Gujarati was trained on only 12M.

\subsection{Training models on this dataset}
We assess the quality of the resulting dataset by learning unsupervised word and sentence representations through fastText and BERT models. For fastText, we train 300-dimensional word embeddings on the head, middle and tail subsets of the English and Polish CommonCrawl corpora, sorted by document perplexity. We evaluate these on standard semantic and syntactic analogy datasets ~\cite{mikolov2013distributed}. We observe in Table~\ref{tab:fasttext} a steady increase in performance as we go from the tail to the head of the dataset, confirming the positive impact of our filtering method based on document perplexity.

\begin{table}[h!]
    \begin{center}
        \begin{tabular}[b]{lcccccc}
        \toprule
        & \multicolumn{3}{c}{English} & \multicolumn{3}{c}{Polish} \\
        \cmidrule(lr){2-4} \cmidrule(lr){5-7} 
               & Total & Sem & Syn  & Total & Sem & Syn \\
        \midrule
        head   & 77.9 & 81.2 & 75.3 & 65.3 & 66.5 & 64.1 \\
        mid.   & 74.2 & 79.0 & 70.4 & 62.8 & 62.7 & 63.0 \\
        tail   & 62.0 & 68.1 & 57.3 & 59.9 & 59.8 & 60.1 \\
        \bottomrule
        \end{tabular}
        \caption{Impact of corpus quality on the quality of fastText word embeddings. We evaluate on semantic and syntactic similarity datasets.\label{tab:fasttext}}
    \end{center}
\end{table}


We also train BERT models on the English (en), Russian (ru), Chinese (zh) and Urdu (ur) languages, using either the Wikipedia corpora or our new CommonCrawl datasets. For these languages, we use respectively 16G, 5G, 1.1G and 106M of raw Wikipedia data (full datasets), and we cap the head CommonCrawl data to 21G, 21G, 17G, 2.2G for English, Russian, Chinese and Urdu. That is, we consider roughly the same amount of data for English, but increase the amount of data for Russian, Chinese and Urdu. We train a BERT-BASE architecture \cite{devlin2018bert} on each of these corpora, without next sentence prediction (NSP) as in ~\cite{lample2019cross}. For better comparison, we early-stop all our models after two days of training on 16 Volta32 GPUs, and use the exact same number of steps for each model. We evaluate each model on the XNLI~\cite{conneau2018xnli} corpus by using the training data in each language. Results presented in Table~\ref{tab:bert} indicate that BERT-BASE models trained on CommonCrawl outperform identical models trained on Wikipedia by 3.3\% on average. With the same amount of data for English, the BERT-BASE model trained on our corpus outperforms the one trained on the Wikipedia. For low-resource languages like Urdu (ur), the Wikipedia dataset being too small, the model pretrained on Wikipedia obtains similar performance than a randomly initialized model. Using our corpus instead, we obtain a 7 points improvement in accuracy, which demonstrates how our filtered corpus can enable language model pretraining for low-resource languages.

\begin{table}[h!]
    \begin{center}
        \begin{tabular}[b]{l|ccccc}
        \toprule
               & en & ru & zh & ur & $\Delta$  \\
        \midrule
        Wiki   & 82.8 & 73.3 & 77.0 & 57.3 & 72.6 \\
        CC     & 85.0 & 76.4 & 77.9 & 64.3 & 75.9 \\
        \bottomrule
        \end{tabular}
        \caption{XNLI dev accuracy for English, Russian, Chinese and Urdu ($\Delta$ for average) for BERT-BASE models trained either on Wikipedia or CommonCrawl. The additional data provided by our pipeline alleviates the lack of resources in most languages and enables representation learning for low-resource languages such as Urdu.\label{tab:bert}}
    \end{center}
\end{table}










\section{Conclusion}

In this paper, we present a pipeline to create curated monolingual corpora in more than $100$ languages.
We preprocess Common Crawl by following the pipeline of~\cite{grave2018learning}, with the differences that we preserve the structure of documents and filter the data based on their distance to Wikipedia.
This improves the quality of the resulting dataset and allows for the training of multilingual text level representations like XLM~\cite{lample2019cross}.


\section*{References}
\bibliographystyle{lrec}
\bibliography{main}

\appendix

\clearpage
\onecolumn


\begin{center}
    \sisetup{
        scientific-notation=engineering,
        round-mode=places,  
        round-precision=3
    }
    \csvreader[
        longtable={c*{4}{S[table-format=3.3e2]}},
        table head={\toprule Language &\multicolumn{1}{c}{Documents} &\multicolumn{1}{c}{Sentences} &\multicolumn{1}{c}{Tokens} &\multicolumn{1}{c}{Size in bytes} \\\midrule},
        table foot={
            \bottomrule
            \caption{
            Number of documents, sentences and tokens after deduplication.
            }},
        filter ifthen={\csvcolii > 999\textbf{}}
    ]%
    {stats_2019-09_all.csv}{}{\csvcoli & \csvcolii  & \csvcoliii & \csvcoliv & \csvcolv }
\label{table:table_counts_all}
\end{center}

\label{main:ref}


\end{document}